\documentclass{article}
\usepackage{ijcai26}

\usepackage{times}
\usepackage{soul}
\usepackage{url}
\usepackage[hidelinks]{hyperref}
\usepackage[utf8]{inputenc}
\usepackage[small]{caption}
\usepackage{graphicx}
\usepackage{amsmath}
\usepackage{amssymb}
\usepackage{amsthm}
\usepackage{booktabs}
\usepackage{multirow}
\usepackage{algorithm}
\usepackage{algorithmic}
\usepackage[switch]{lineno}
\usepackage{xcolor}
\usepackage{comment}

\newtheorem{definition}{Definition}

\newcommand{\equalcontrib}{\textsuperscript{\dag}}
\title{Beyond Heatmaps: Unsupervised Concept-Graph Reasoning for Interpretable Visual Explanation}

\author{
Md Mohasin Hossain$^{1,2}$\equalcontrib
\and
Anar Amirli$^{4}$\equalcontrib\thanks{Work performed while affiliated with Saarland University.}
\and 
Robert Leist$^{1}$\and 
Md Abdul Kadir$^{1,3}$\And
Daniel Sonntag$^{1,3}$\\
\affiliations
$^1$German Research Center for Artificial Intelligence (DFKI), Saarbrücken, Germany\\
$^2$Saarland University, Saarbrücken, Germany\\
$^3$Oldenburg University, Oldenburg, Germany\\
$^4$BEGO GmbH \& Co. KG, Bremen, Germany\\
\emails
}
\begin{document}

\maketitle

\begingroup
\renewcommand{\thefootnote}{\fnsymbol{footnote}}
\footnotetext[2]{Equal contribution.}
\endgroup

% Input the sections of the paper
\begin{abstract}
Concept Bottleneck Models (CBMs) provide an intrinsically interpretable alternative to post-hoc explanations. However, existing CBMs often rely on predefined concept vocabularies or supervised annotations, lack explicit concept grounding, and summarize each concept with a single image-level score---discarding spatial recurrence and inter-concept dependencies. We propose a Graph-based Concept Bottleneck Model (G-CBM), an intrinsically interpretable framework that performs unsupervised concept discovery via Non-negative Matrix Factorization (NMF) and represents the discovered concepts as nodes in a per-image concept-graph representation. G-CBM matches region-level features to these concept nodes---providing concept grounding and capturing concept recurrence across the image---and applies a \emph{tunable concept filtering threshold} $\tau$ to suppress weak region-level features. A Graph Attention Network (GAT) then performs concept-level reasoning by modeling nonlinear dependencies across nodes. Across ImageNet, HAM10000, PH2, and Derm7pt, G-CBM achieves an average relative AUC improvement of 3.7\% over a ResNet-50 baseline. Concept filtering frequently improves predictive performance while inducing selective concept use, achieving peak AUC of $0.96$ on PH2 with only 2 of 10 concepts and 0.92 on HAM10000 with 3.8 of 9 concepts. On dermoscopy benchmarks, G-CBM is competitive with supervised approaches requiring external annotations. Deletion/insertion analyses with random ablation controls show that the learned concept ranking faithfully reflects model predictions.
\end{abstract}

\vspace{-15pt}
\section{Introduction}
\begin{figure}[t]
    \centering
    \includegraphics[width=\linewidth]{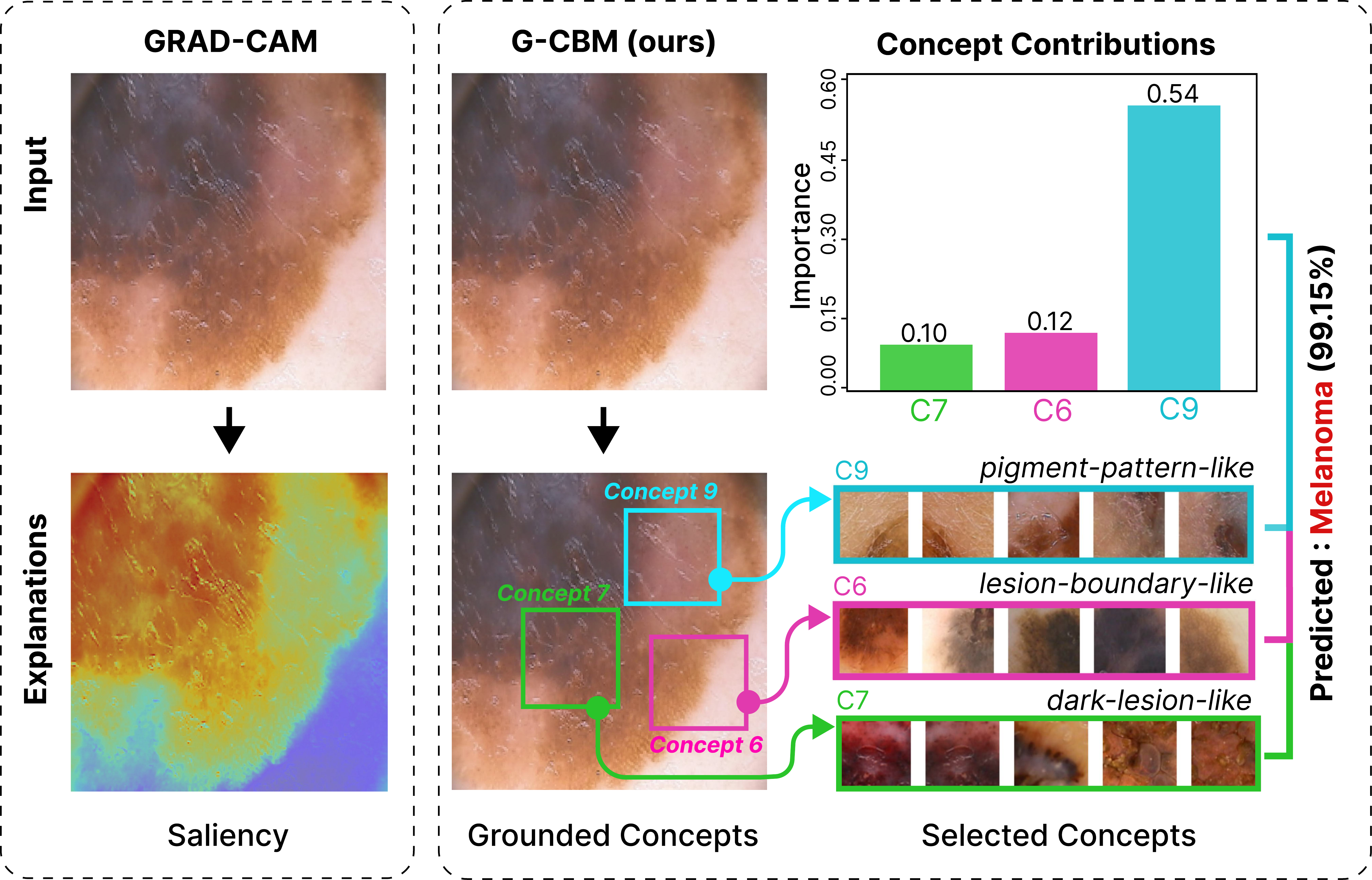}
    \caption{G-CBM produces an interpretable melanoma diagnosis (99.15\%) on a dermoscopy image. \textit{Left}: Grad-CAM yields a diffuse pixel-level saliency map. \textit{Right}: G-CBM jointly provides \emph{concept selection} (\emph{what} concepts drive the prediction: \textit{pigment-pattern-like}, \textit{lesion-boundary-like}, \textit{dark-lesion-like}), \emph{concept grounding} (\emph{where} they appear, shown as bounding boxes), and \emph{importance scores} (\emph{how much} each contributes, shown as a bar chart).}
    \label{fig:teaser}
     \vspace{-15pt} 
\end{figure}

\vspace{-3pt}
Deep learning models have achieved remarkable success across computer vision tasks~\cite{He2016,Dosovitskiy2021}, yet their opaque decision processes limit deployment in safety-critical domains~\cite{Esteva2019}. A common remedy is to explain predictions using post-hoc saliency-based methods such as Grad-CAM~\cite{Selvaraju2017} or Integrated Gradients~\cite{Sundararajan2017}. Although these methods can highlight discriminative image regions, they operate at the pixel level and provide limited insight into the visual concepts influencing a prediction, and several fail basic sanity checks~\cite{Adebayo2018}. As illustrated in Fig.~\ref{fig:teaser} (left), Grad-CAM produces only a diffuse, unstructured saliency map that highlights image regions without identifying \emph{what} they represent. Post-hoc concept-based methods offer a more structured alternative to these saliency methods by explaining predictions through intermediate concepts rather than raw pixels~\cite{fel2023craftconceptrecursiveactivation,patrício2023coherentconceptbasedexplanationsmedical}. However, these methods do not perform concept-level reasoning within the prediction process.

Concept Bottleneck Models (CBMs)~\cite{Koh2020} integrate
interpretability through concept-level reasoning within the prediction
process, but still suffer from several limitations. Existing CBMs
typically rely on predefined concept vocabularies or supervised concept
annotations~\cite{Oikarinen2023,Yuksekgonul2022}, which are expensive to
obtain or outsource interpretability to opaque foundation models such
as LLMs and CLIP. While the linear classifier weights provide coarse
per-concept attributions (\emph{how much}), they cannot model nonlinear
inter-concept dependencies and reduce each concept to a single
image-level score, thereby ignoring its recurrence across image regions.
Moreover, CBMs lack explicit concept grounding in the input space
(\emph{where}), making it difficult to verify whether the predicted
concepts correspond to meaningful visual evidence.

Taken together, post-hoc saliency methods answer only \emph{where}; post-hoc concept methods answer \emph{what} concepts are present but do not integrate concepts into the prediction process; and CBMs integrate interpretability into the prediction process and answer \emph{what} and \emph{how much}, but rely on costly supervision or predefined vocabularies, provide weak concept grounding (no \emph{where}), and combine concepts through a linear classifier---ignoring nonlinear inter-concept dependencies.

We propose \textbf{Graph-based Concept Bottleneck Models (G-CBM)}, an intrinsically interpretable image classification framework that integrates concept-level reasoning directly into the prediction pipeline.  G-CBM extracts region-level features by passing image patches through a frozen backbone. G-CBM discovers visual concepts from these intermediate features using Non-negative Matrix Factorization (NMF)\cite{fel2023craftconceptrecursiveactivation}, yielding reusable concept bases that capture recurring visual patterns. It constructs a per-image concept-graph representation: a graph with one \emph{concept node} per discovered concept basis, where a concept node aggregates the region-level features of image patches that match its corresponding basis, and a Graph Attention Network (GAT) models nonlinear inter-concept dependencies across all concept nodes. By matching region-level features against concept nodes, G-CBM preserves repeated and spatially distributed concept evidence across the image. This produces three interpretive signals (Fig.~\ref{fig:teaser}, right): activated concept nodes perform \emph{concept selection} by indicating \emph{what} concepts influence the prediction, the image regions provide \emph{concept grounding} showing \emph{where} the evidence appears, and gradient-based sensitivities with respect to concept nodes serve as 
\emph{importance scores} quantifying \emph{how much} each concept contributes.

In summary, our contributions are as follows. First, we propose G-CBM, an intrinsically interpretable framework that jointly performs classification and explanation (Fig.~\ref{fig:pipeline}). G-CBM combines \emph{unsupervised concept discovery} with graph-based relational reasoning to produce \emph{concept selection}, \emph{concept grounding}, and \emph{importance scores} in a unified pipeline. Second, we introduce a per-image concept-graph representation in which nodes aggregate region-level features and edges, learned via graph attention, capture inter-concept dependencies (Sec.~\ref{sec:graph}). Third, we introduce a \emph{tunable concept filtering threshold} $\tau$ that suppresses 
weak region-level features prior to node aggregation, enabling explicit control over explanation selectivity while improving concept grounding 
(Sec.~\ref{sec:graph}, Table~\ref{tab:budget}); on PH2, this yields an AUC of 0.96 using only 2 of 10 available concepts. Finally, we validate G-CBM across four datasets  (ImageNet, HAM10000, PH2, Derm7pt), demonstrating an average relative AUC improvement of 3.7\% over ResNet-50, competitive performance against supervised dermoscopy CBMs without requiring manual concept annotations, and faithful concept rankings under deletion/insertion analyses with random ablation controls (Tables~\ref{tab:perf},~\ref{tab:sota},~\ref{tab:fidelity}).

\vspace{-10pt}
\section{Related Work}
\vspace{-3pt}
Our review covers four areas: post-hoc saliency-based methods, post-hoc concept-based methods, graph-based explanation methods, and Concept Bottleneck Models (CBMs).

Post-hoc saliency-based methods produce pixel-level maps highlighting discriminative regions. Grad-CAM~\cite{Selvaraju2017}, Integrated Gradients~\cite{Sundararajan2017}, and SmoothGrad~\cite{Smilkov2017} indicate \emph{where} a model attends but not the underlying visual concepts~\cite{nunnari2021overlap}. Adebayo~et~al.~\shortcite{Adebayo2018} and Kindermans~et~al.~\shortcite{Kindermans2019} argue that many saliency methods lack reliability and fail sanity checks, motivating research into more structured and semantically meaningful explanations.

Post-hoc concept-based methods move beyond pixels by identifying visual concepts (\emph{what}) in intermediate-layer activations. TCAV~\cite{Kim2018} measures sensitivity to user-defined concept directions. ACE~\cite{Ghorbani2019} discovers concepts by clustering superpixel activations, while CRAFT~\cite{fel2023craftconceptrecursiveactivation} extracts features from image regions and applies Non-negative Matrix Factorization (NMF), yielding more interpretable, part-based bases than PCA and K-Means---motivating our use of NMF in G-CBM. Concept Relevance Propagation (CRP)~\cite{achtibat2023attribution} goes furthest among these, conditioning the relevance backward pass on concept-encoding channels to localize concepts in the input while quantifying their contribution. Yet all these methods remain post-hoc: they do not integrate concept discovery or inter-concept dependencies into the prediction pipeline, and none model nonlinear inter-concept dependencies.

Concept Bottleneck Models (CBMs)~\cite{Koh2020} integrate interpretability directly into the prediction pipeline by first identifying concepts and then using them for final classification. This explains \emph{what} concepts contribute to a prediction, with linear classifier weights serving as coarse per-concept attributions (\emph{how much}). However, CBMs typically rely on predefined concept vocabularies or supervised concept annotations~\cite{tarek2026traceconceptbottleneckmodel}. Several extensions address these limitations. Label-Free CBMs~\cite{Oikarinen2023} reduce manual annotation by using Large Language Models (LLMs) to generate candidate concepts and CLIP similarity to score their presence. Similar LLM/VLM-based concept bottlenecks have also been used for radiology report generation~\cite{alam2025towards,alam2025cbm}. Causally Structured CBMs~\cite{defelice2025causally} incorporate predefined causal relationships among concepts to improve reasoning consistency.

Dermoscopy-specific variants, PCBM~\cite{Yuksekgonul2022}, CBE~\cite{patrício2023coherentconceptbasedexplanationsmedical}, and MICA~\cite{bie2024micaexplainableskinlesion} combine concept bottlenecks with vision-language representations for clinically interpretable diagnosis. However, these methods inherit two common drawbacks: they rely on predefined or expert-guided concept vocabularies (or on opaque LLMs and CLIP encoders to generate and match them), thereby outsourcing interpretability to non-interpretable foundation models. They typically retain only global (image-level) scalar concept representations, thereby discarding spatial recurrence; do not explicitly model nonlinear dependencies between concepts and ground \emph{where} each concept appears in the image.

Graph-based methods offer a complementary way to model relational structure. GNNExplainer~\cite{Ying2019} and PGExplainer~\cite{Luo2020} identify nodes, edges, or subgraphs for explaining predictions, but assume graph-structured inputs and act as post-hoc explainers for trained GNNs. Scene-graph models~\cite{Johnson2018} perform visual reasoning over object relationships but build graphs from detected objects rather than discovered visual concepts. Consequently, existing graph-based approaches do not yield an intrinsically interpretable concept bottleneck framework.

G-CBM addresses these limitations by integrating \emph{unsupervised concept discovery} and graph-based relational reasoning. G-CBM identifies \emph{what} concepts are present by discovering reusable visual concepts from intermediate-layer region features using NMF, then constructs a per-image concept-graph representation in which each node aggregates region-level features matching a discovered concept, preserving spatial recurrence, and a GAT models inter-concept dependencies. Matching region-level features against concept nodes identifies \emph{where} the concepts appear in the input through concept grounding, and gradient-based class-probability sensitivities of concept nodes quantify \emph{how much} each concept contributes. To our knowledge, G-CBM is the first intrinsically interpretable framework combining unsupervised visual concept discovery with graph-attention-based modeling of inter-concept dependencies.
\vspace{-0.7mm}
\vspace{-5pt}
\section{Methodology}
\vspace{-3pt}
\begin{figure*}[t]
    \centering
    \includegraphics[width=\textwidth]{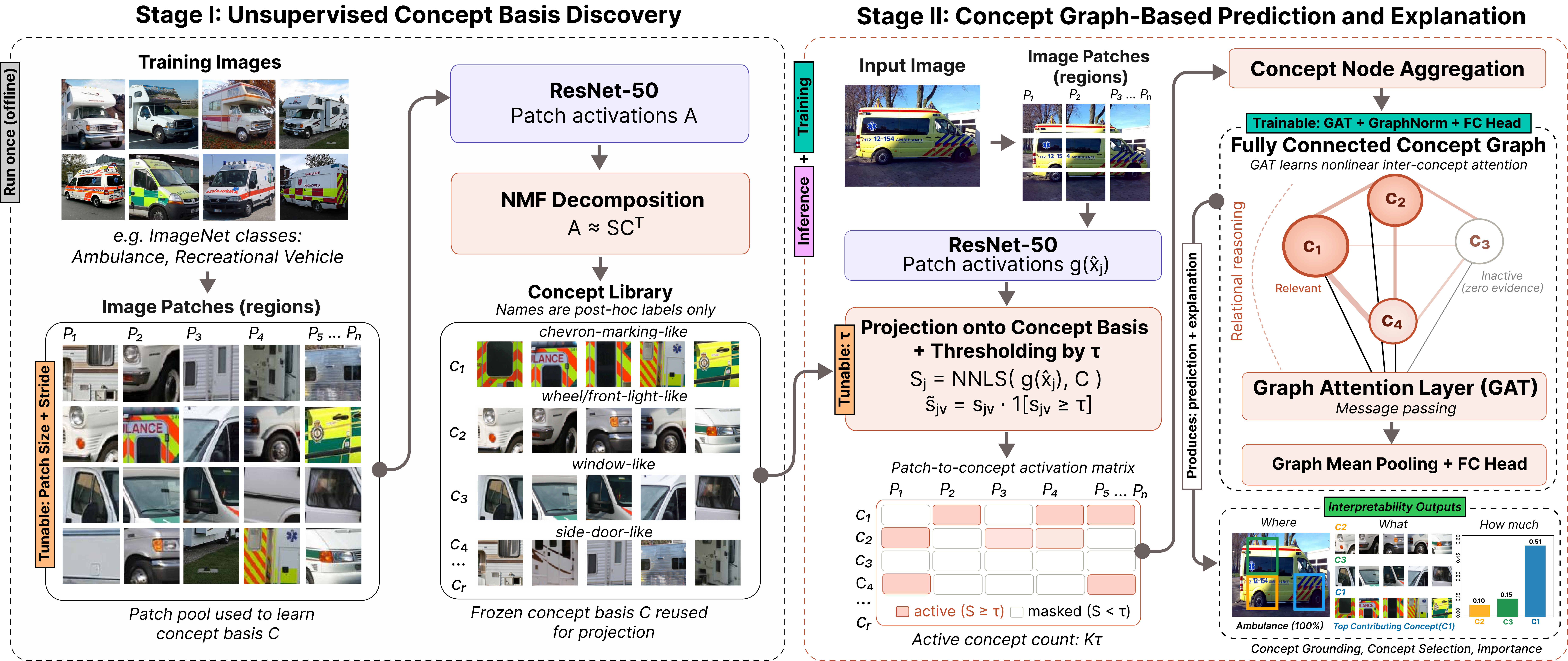}
    \caption{Overview of G-CBM. Stage~I: training images are divided into
    patches (regions) and encoded by a frozen backbone; NMF decomposes the
    activations to obtain a reusable concept basis $\mathbf{C}$. Stage~II: For an input image, region-level features (patch activations) are projected onto $\mathbf{C}$, filtered by the concept filtering threshold $\tau$, and aggregated into concept-specific nodes. A GAT processes the resulting graph
    for prediction and explanation. Concept names
    are post-hoc descriptions for visualization only and are not used during training.}
    \label{fig:pipeline}
    \vspace{-2mm}
\end{figure*}

Our G-CBM framework performs intrinsically interpretable image classification through 
concept-level reasoning over a concept-graph representation. As shown in 
Fig.~\ref{fig:pipeline}, G-CBM first discovers reusable concept 
bases from region-level features extracted from image patches via 
Non-negative Matrix Factorization (NMF), and then classifies each image 
using a Graph Attention Network (GAT).
\vspace{-5pt}

\subsection{Stage I: Unsupervised Concept Basis Discovery}
\label{sec:concept-basis}
We extract patches of size $h\times w$ with stride ratio $\rho$ from 
the training images $\mathcal{X}$, forming a candidate pool 
$\mathbf{X}\in\mathbb{R}^{N\times c\times h\times w}$ with $N$ patches 
in total. Each patch is passed independently through a frozen 
intermediate layer $g(\cdot)$ of a pretrained backbone, yielding 
region-level features stacked into an activation matrix
\vspace{-5pt}
\begin{equation}
\mathbf{A}=g(\mathbf{X})\in\mathbb{R}^{N\times p_{\mathrm{feat}}},
\end{equation}
where $p_{\mathrm{feat}}$ is the hidden feature dimension and each row 
$\mathbf{A}_i^{\top}\in\mathbb{R}^{p_{\mathrm{feat}}}$ is the 
region-level feature of the $i$-th patch.

Following CRAFT~\cite{fel2023craftconceptrecursiveactivation}, we use 
NMF~\cite{Lee1999} for unsupervised concept discovery from $\mathbf{A}$. NMF finds non-negative vector bases in activation space, where each direction corresponds to a distinct piece of information (e.g., wheel, stripe, glass), and the non-negative coefficient of a sample along a direction tells how strongly that concept is present.
\vspace{-5pt}
\begin{equation}
\mathbf{S}^{*},\mathbf{C}^{*}\in
\arg\min_{\mathbf{S}\geq 0,\mathbf{C}\geq 0}
\tfrac{1}{2}\|\mathbf{A}-\mathbf{S}\mathbf{C}^{\top}\|_F^{2}.
\end{equation}
Here, $\mathbf{C}\in\mathbb{R}^{p_{\mathrm{feat}}\times r}$ holds the 
$r$ concept bases as columns, while $\mathbf{S}\in\mathbb{R}^{N\times r}$ contains the region-level concept scores used only within Stage~I for concept-number selection.

\vspace{-5pt}
\begin{equation}
    \mathbf{A}_i \approx \sum_{v=1}^{r}s_{iv}\,\mathbf{c}_v,
    \label{eq:nmf-reconstruction}
\end{equation}
where $\mathbf{c}_v\in\mathbb{R}^{p_{\mathrm{feat}}}$ is the $v$-th 
column of $\mathbf{C}$. The selected concept bases $\mathbf{C}$ are 
then frozen and reused as the concept nodes (each concept basis $\mathbf{c}_v$ defines a concept node $v \in \mathcal{V}$)  in our per-image 
concept-graph representation (Sec.~\ref{sec:graph}).

\paragraph{Selecting $r$.}
For each $k\in\{6,\ldots,16\}$ we score NMF as
$\mathrm{Score}(k)=\tfrac{1}{k}\sum_i D_i-\lambda P_k$, where $D_i$ is concept $i$'s top-$10\%$ patch fraction in its dominant class and $P_k$
penalises class-imbalanced coverage among discriminative concepts ($D_i\geq 0.6$).
We set $\lambda=1$ and pick $r=\arg\max_k \mathrm{Score}(k)$ per
(backbone, dataset); selected values in
Table~\ref{tab:threshold_graph_summary}.

\subsection{Stage II: Concept Graph-Based Prediction and Explanation}
\label{sec:graph}
\paragraph{Projection onto concept basis.}
For any image $x$, its $n$ patches 
$\hat{\mathbf{X}}$ are passed through the backbone $g(\cdot)$, 
where $n$ denotes the per-image patch count (distinct from 
$N$, the total training patch pool used in Sec.~\ref{sec:concept-basis}).
The resulting region-level features are then projected onto the frozen 
concept bases $\mathbf{C}$ via Non-negative Least Squares (NNLS):
\begin{equation}
    \mathbf{s}_j^{*}\in
    \arg\min_{\mathbf{s}\geq 0}
    \tfrac{1}{2}\|g(\hat{X}_j)-\mathbf{C}\mathbf{s}\|_2^{2},
    \label{eq:nnls}
\end{equation}
where $\hat{X}_j\in\mathbb{R}^{c\times h\times w}$ is the $j$-th patch, 
$g(\hat{X}_j)\in\mathbb{R}^{p_{\mathrm{feat}}}$ is its region-level 
feature, and $\mathbf{s}_j=[s_{j1},\ldots,s_{jr}]^{\top}\in\mathbb{R}^{r}$ 
contains its region-level concept scores. We build a fully connected 
concept graph 
$\mathcal{G}=(\mathcal{V},\mathcal{E},\mathbf{H})$ with node set 
$\mathcal{V}$ of $r$ concept nodes, edge set $\mathcal{E}$, and node 
features $\mathbf{H}\in\mathbb{R}^{r\times p_{\mathrm{feat}}}$, 
preserving repeated and spatially distributed concept evidence across 
the image.

\paragraph{Concept filtering threshold.}
Many patches bear only marginal resemblance to any concept, contributing 
noise to the concept node representations. We therefore introduce a 
concept filtering threshold $\tau\geq 0$:
\begin{equation}
    \tilde{s}_{jv}=s_{jv}\,\mathbf{1}[s_{jv}\geq\tau].
    \label{eq:threshold}
\end{equation}
$\tau \geq 0 $ filters patches with marginal concept evidence, controlling explanation selectivity and improving grounding without removing nodes from the fixed graph. At  $\tau = 0 $, G-CBM reduces to the unthresholded model; if all nodes become inactive at high $\tau$, prediction confidence degrades toward chance. We treat $\tau$ as a tunable hyperparameter and select $\tau^*$ per (backbone, dataset) on the validation split
(Sec.~\ref{sec:budget}).
\begin{definition}[Active Concept Count]
\label{def:active-count}
For an image $x$, a concept node $v$ is \emph{active after thresholding} 
if at least one patch contributes positive thresholded evidence, i.e., 
$\sum_{j=1}^{n}\tilde{s}_{jv}>0$. The active concept count $K_\tau(x)$ 
is the number of such nodes:
\begin{equation}
    K_{\tau}(x)=
    \sum_{v=1}^{r}
    \mathbf{1}\!\left[\sum_{j=1}^{n}\tilde{s}_{jv}>0\right].
\end{equation}
\end{definition}

$K_\tau(x)$ measures the explanation selectivity for image $x$: a smaller 
value means fewer concepts contribute to the prediction. We report $\bar{K}_\tau$, the mean of $K_\tau(x)$ over the test set, as a dataset-level selectivity measure (see Fig.~\ref{fig:thresh}).

\paragraph{Node feature initialization.}
Each concept node $v$ is initialized by aggregating the region-level 
features $g(\hat{X}_j)$ weighted by the corresponding thresholded 
concept scores $\tilde{s}_{jv}$:
\vspace{-3pt}
\begin{equation}
    \mathbf{h}_v^{(0)}
    \;=\;
    \mathrm{GELU}\!\left(
        \frac{1}{n}\sum_{j=1}^{n} g(\hat{X}_j)\,\tilde{s}_{jv}
    \right),
    \qquad v=1,\ldots,r,
    \label{eq:node}
\end{equation}
where $\mathrm{GELU}$~\cite{hendrycks2016gelu} is applied element-wise. 
The $1/n$ normalization weights each concept by its activation frequency. Stacking 
$\{\mathbf{h}_v^{(0)}\}_{v=1}^{r}$ row-wise yields the initial node 
feature matrix $\mathbf{H}\in\mathbb{R}^{r\times p_{\mathrm{feat}}}$, 
which is passed to the GAT.
% (Sec.~\ref{sec:gat})

% \subsection{Classification via Graph Attention}
\paragraph{Classification via Graph Attention.}
\label{sec:gat}

The fully connected concept graph is processed with a shallow multi-head
GAT~\cite{velickovic2018gat}, allowing each concept node to attend to all
others. For head $i \in \{1, \ldots, H\}$, with $\mathbf{W}^{(i)}\in\mathbb{R}^{d'\times p_{\mathrm{feat}}}$ 
and learnable attention vector $\mathbf{a}^{(i)}\in\mathbb{R}^{2d'}$, we 
compute the attention energy $e_{vu}^{(i)}$ and coefficient 
$\alpha_{vu}^{(i)}$ from node $u$ to node $v$ as

\vspace{-10pt}
\begin{align}
    e_{vu}^{(i)}
    &=
    \mathrm{LeakyReLU}\!\left(
    \mathbf{a}^{(i)\top}
    [\mathbf{W}^{(i)}\mathbf{h}_v^{(0)}
    \Vert
    \mathbf{W}^{(i)}\mathbf{h}_u^{(0)}]\right), \\
    \alpha_{vu}^{(i)}
    &=
    \frac{\exp(e_{vu}^{(i)})}
    {\sum_{w\in\mathcal{N}(v)}\exp(e_{vw}^{(i)})},
    \label{eq:gat-attn}
\end{align}
where $\mathcal{N}(v)$ contains all concept nodes. The attention-head 
outputs are averaged and then normalized with 
GraphNorm~\cite{cai2021graphnorm} before ELU activation:
\vspace{-6pt}
\begin{equation}
    \bar{\mathbf{m}}_v
    =
    \frac{1}{n_h}
    \sum_{i=1}^{n_h}
    \sum_{u\in\mathcal{N}(v)}
    \alpha_{vu}^{(i)}
    \mathbf{W}^{(i)}\mathbf{h}_u^{(0)} ,
    \label{eq:gat-message}
\end{equation}
\vspace{-8pt}
\begin{equation}
    \mathbf{h}_v^{(1)}
    =
    \mathrm{ELU}\!\left(
    \mathrm{GraphNorm}(\bar{\mathbf{m}}_v)
    \right).
    \label{eq:gat}
\end{equation}
The GAT models nonlinear inter-concept dependencies and applies
node-specific transformations, so each concept node makes a distinguishable
contribution to prediction (Eq.~\ref{eq:concept-score}); under a linear
or MLP head on pooled features, all nodes would contribute identically,
collapsing importance scores.
\vspace{-10pt}

\begin{equation}
    \tilde{\mathbf{h}}=\frac{1}{r}\sum_{v=1}^{r}\mathbf{h}_v^{(1)},\qquad
    \hat{\mathbf{y}}=
    \mathrm{Softmax}(\mathbf{W}_{\mathrm{cls}}\tilde{\mathbf{h}}
    +\mathbf{b}_{\mathrm{cls}}).
    \label{eq:classify}
\end{equation}
During training, only the GAT and classification head are trained (cross-entropy); the backbone and concept bases remain frozen.

\vspace{-5pt}
\subsection{Interpretability Outputs: Selection, Grounding, and Importance}
\label{sec:explanations}

\paragraph{Concept selection (\emph{what}).}
The set of concept nodes with positive thresholded evidence 
$\{v : \sum_{j=1}^{n}\tilde{s}_{jv}>0\}$, of size $K_\tau(x)$ 
(Def.~\ref{def:active-count}), identifies which concepts influence the 
prediction for image $x$.

\paragraph{Importance scores (\emph{how much}).}
Let $c^*=\arg\max_c \hat{y}_c$ be the predicted class. The importance of
a concept node $v$ for class $c^*$ is measured by the gradient-based
sensitivity of the predicted-class probability to the node feature:
\vspace{-5pt}
\begin{equation}
    S_C(v,c^*)=
    \left\|
    \frac{\partial \hat{y}_{c^*}}
    {\partial \mathbf{h}_v^{(0)}}
    \right\|_1 .
    \label{eq:concept-score}
\end{equation}
Ranking concept nodes by $S_C$ quantifies how much each concept
contributes to the final decision.

\paragraph{Concept grounding (\emph{where}).}
For image $x$, we identify the most influential concepts by ranking
all nodes according to $S_C$ and retaining the top three concepts. To
localise each concept in the image, we assign every patch
$\hat{X}_j$ a grounding score based on how strongly it activates
concepts that contribute to the final prediction:
\vspace{-8pt}
\begin{equation}
    S_P(j,c^*)
    =
    \sum_{v=1}^{r}
    \tilde{s}_{jv}\, S_C(v,c^*) .
    \label{eq:patch-score}
\end{equation}
We treat $\tilde{s}_{jv}$ as constants since NNLS is non-differentiable.
A patch scores high when it strongly activates concepts important to the
decision; for each selected concept, we visualize its top-activating
patches  (Fig.~\ref{fig:spatial}).

\vspace{-10pt}
\section{Experiments and Results}
We evaluate G-CBM in terms of classification performance, faithfulness, and spatial interpretability across natural and dermoscopic image datasets.

\subsection{Datasets and Setup}
\label{sec:setup}

Our experiments span four datasets. An \textbf{ImageNet} subset~\cite{Deng2009} (2,943 images) evaluates fine-grained differentiation between visually similar categories (Ambulance vs.\ Recreational Vehicle). The remaining dermoscopy datasets focus on binary melanoma vs.\ nevus classification: \textbf{HAM10000}~\cite{Tschandl2018HAM10000} ($\sim$7,000 images), \textbf{PH2}~\cite{Mendonca2013} (200 images), and \textbf{Derm7pt}~\cite{Kawahara2018Derm7pt} (827 images). All datasets use a 70/15/15 train/val/test split, with minority-class oversampling applied only to the training split.

\paragraph{Implementation.}
We use ResNet-50~\cite{He2016}, DenseNet-201~\cite{Huang2017}, and MobileNet-V2~\cite{Sandler2018} in our experiments. Region-level features come from the last conv layer, patch size $70\times70$ with a stride ratio $\rho=0.5$. The GAT uses a hidden dimension of 128 with four attention heads for PH2 and Derm7pt, and six for HAM10000 and ImageNet. Models are trained for up to 300 epochs using AdamW (lr$=10^{-3}$, wd$=2\times10^{-4}$) with early stopping on validation loss. Results are reported as mean\,$\pm$\,std over three runs with seeds $\{42,123,456\}$.

\paragraph{Concept Discovery.}

To verify NMF transfers to dermoscopy, we replicate the PCA/K-Means/NMF comparison across all three
backbones on ImageNet and HAM10000 (Table~\ref{tab:nmf}). NMF provides
a favorable trade-off across reconstruction error, sparsity,
stability, and OOD robustness in both domains, confirming its
suitability for G-CBM.

\begin{table}[t]
    \resizebox{\columnwidth}{!}{%
    \begin{tabular}{llrrrr}
        \toprule
        Backbone & Method & $\ell_2$ $\downarrow$ & Spars. $\uparrow$ & Stab. $\downarrow$ & OOD $\downarrow$ \\
        \midrule
        \multirow{3}{*}{ResNet-50}
          & PCA     & 0.38/0.27 & 0.00/0.00 & 0.28/0.37 & 0.15/0.06 \\
          & K-Means & 0.50/0.37 & 0.96/0.96 & 0.01/0.02 & 0.14/0.05 \\
          & NMF     & 0.40/0.28 & 0.36/0.35 & 0.09/0.07 & 0.16/0.06 \\
        \midrule
        \multirow{3}{*}{DenseNet-201}
          & PCA     & 0.59/0.41 & 0.00/0.00 & 0.33/0.33 & 0.34/0.13 \\
          & K-Means & 0.66/0.50 & 0.96/0.96 & 0.04/0.04 & 0.28/0.10 \\
          & NMF     & 0.58/0.40 & 0.41/0.34 & 0.14/0.09 & 0.32/0.11 \\
        \midrule
        \multirow{3}{*}{MobileNet-V2}
          & PCA     & 0.43/0.30 & 0.00/0.00 & 0.36/0.39 & 0.19/0.07 \\
          & K-Means & 0.54/0.41 & 0.96/0.96 & 0.03/0.03 & 0.17/0.05 \\
          & NMF     & 0.46/0.32 & 0.41/0.40 & 0.12/0.11 & 0.20/0.07 \\
        \bottomrule
    \end{tabular}}
    \caption{Quality of discovered concepts across backbones and methods (ImageNet/HAM10000 per cell). Metrics follow: $\ell_2$ reconstruction error, sparsity, stability, and OOD error. NMF achieves the best overall balance and is used in G-CBM.}
    \label{tab:nmf}
    \vspace{-3mm}
\end{table}

\begin{figure}[!t]
    \centering
    \includegraphics[width=\linewidth]{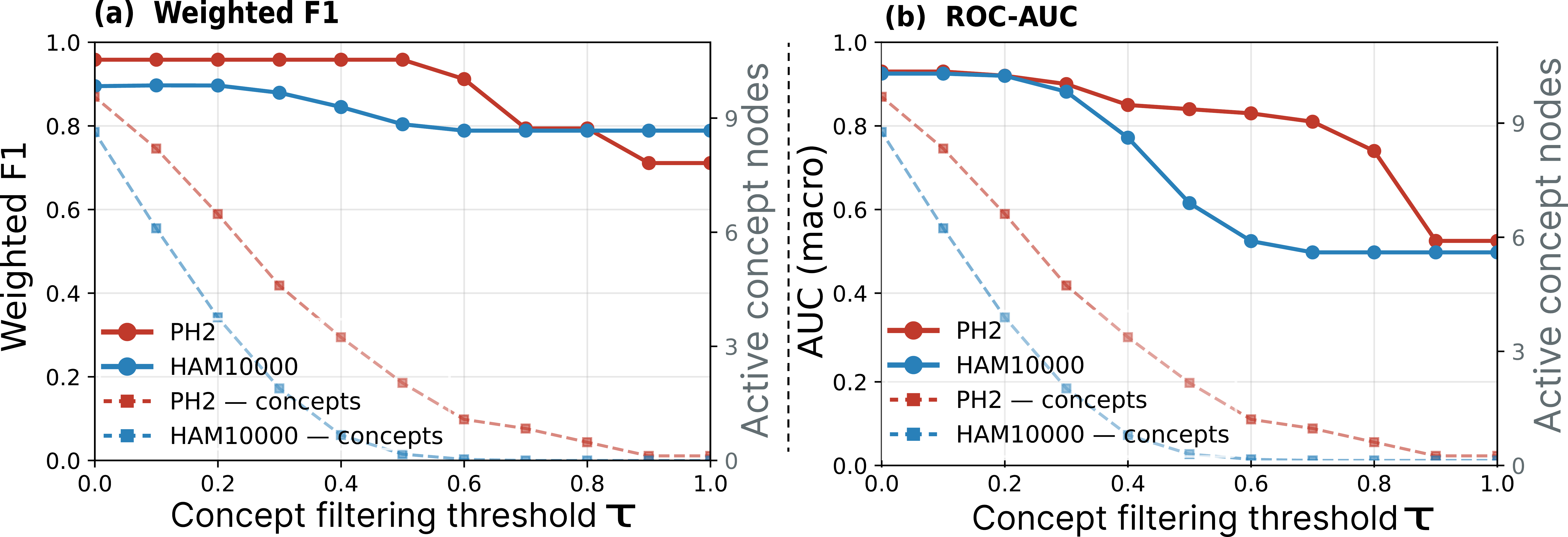}
    \caption{Concept selectivity vs.\ classification performance on PH2 and
    HAM10000. \textit{Left}: weighted F1 (solid) and mean active concept
    count $\bar{K}_\tau$ (dashed) vs.\ $\tau$. \textit{Right}: AUC (solid)
    and $\bar{K}_\tau$ (dashed) vs.\ $\tau$. F1 remains elevated at high
    $\tau$ due to class imbalance; AUC better reflects threshold-dependent
    discriminative quality.}
    \label{fig:thresh}
    \vspace{-3mm}
\end{figure}

\paragraph{Concept Filtering.}
\label{sec:budget}
Since tuning $\tau^*$ across datasets and backbones is expensive, we calibrate it on validation: each model is trained once at $\tau=0$,
$\tau\in\{0.1,\ldots,1.0\}$ is applied at  validation, and $\tau^*$ is selected
by validation F1 score (Table~\ref{tab:threshold_graph_summary}). Fig.~\ref{fig:thresh}
plots F1, AUC, and $\bar{K}_\tau$ versus $\tau$ for PH2 and HAM10000: for
small $\tau$, F1 and AUC do not drop significantly, indicating that mild
filtering removes noise without sacrificing discriminative information. Models are then retrained at $\tau^*$.
Calibration statistics differ from test because filtering shifts the GAT's input distribution (e.g., PH2/ResNet-50: 0.833 vs.\ 0.960 after retraining). Table~\ref{tab:threshold_graph_summary} reports the selected $\tau^*$ and
the corresponding validation statistics for all backbones and datasets. After selecting $\tau^*$, we retrain, validate, and
test each model with $\tau^*$ applied throughout; final test-set results
are reported in Section~\ref{sec:perf}.

\vspace{-6pt}
\begin{table}[ht]
\centering
\scriptsize
\setlength{\tabcolsep}{3pt}
\renewcommand{\arraystretch}{1.05}
\resizebox{\columnwidth}{!}{%
\begin{tabular}{lccc}
\toprule
& ResNet-50 & DenseNet-201 & MobileNet-V2 \\
\cmidrule(lr){2-2}\cmidrule(lr){3-3}\cmidrule(lr){4-4}
\textbf{Dataset} &
$\tau^*$/F1/AUC/$\bar{K}_{\tau^*}$/$r$ &
$\tau^*$/F1/AUC/$\bar{K}_{\tau^*}$/$r$ &
$\tau^*$/F1/AUC/$\bar{K}_{\tau^*}$/$r$ \\
\midrule
HAM10000 &
\textbf{0.2}/0.894/0.917/3.76/9 &
\textbf{0.3}/0.880/0.892/3.74/16 &
\textbf{0.2}/0.884/0.907/2.47/6 \\
PH2 &
\textbf{0.5}/0.958/0.833/2.04/10 &
\textbf{0.3}/0.933/0.980/6.08/12 &
\textbf{0.4}/0.922/0.967/2.28/8 \\
Derm7pt &
\textbf{0.1}/0.821/0.885/9.07/12 &
\textbf{0.1}/0.804/0.852/6.25/7 &
\textbf{0.1}/0.797/0.832/5.31/7 \\
ImageNet &
\textbf{0.1}/0.949/0.989/6.31/8 &
\textbf{0.1}/0.951/0.994/11.30/16 &
\textbf{0.2}/0.913/0.977/5.10/10 \\
\bottomrule
\end{tabular}%
}
\caption{Selected thresholds $\tau^*$ and corresponding
\emph{validation} statistics per backbone and dataset, used only for
$\tau^*$ selection (no test-set numbers here).
$r$: total concept nodes; $\bar{K}_{\tau^*}$: mean active nodes at
$\tau^*$. Each cell lists $\tau^*$/F1/AUC/$\bar{K}_{\tau^*}$/$r$.
Means over three seeds. Test-set results obtained after retraining at
$\tau^*$ are reported in Table~\ref{tab:perf}.}
\label{tab:threshold_graph_summary}

\end{table}
\begin{table*}[!ht]
    \resizebox{\textwidth}{!}{%
    \begin{tabular}{lrrrrrrrr}
        \toprule
        & \multicolumn{2}{c}{HAM10000} & \multicolumn{2}{c}{PH2}
        & \multicolumn{2}{c}{Derm7pt} & \multicolumn{2}{c}{ImageNet} \\
        \cmidrule(lr){2-3}\cmidrule(lr){4-5}\cmidrule(lr){6-7}\cmidrule(lr){8-9}
        Model & AUC & F1 & AUC & F1 & AUC & F1 & AUC & F1 \\
        \midrule
        \multicolumn{9}{l}{\textit{CNN baselines (no concept bottleneck)}} \\
        CNN ResNet-50       & 0.891$\pm$.011 & 0.865$\pm$.019 & 0.903$\pm$.096 & 0.901$\pm$.085 & 0.828$\pm$.029 & 0.791$\pm$.014 & 0.980$\pm$.004 & 0.923$\pm$.006 \\
        CNN DenseNet-201    & 0.901$\pm$.005 & 0.877$\pm$.008 & 0.945$\pm$.015 & 0.864$\pm$.045 & 0.850$\pm$.030 & 0.796$\pm$.006 & 0.973$\pm$.007 & 0.922$\pm$.012 \\
        CNN MobileNet-V2    & 0.912$\pm$.013 & 0.892$\pm$.009 & 0.943$\pm$.005 & 0.908$\pm$.017 & \textbf{0.901}$\pm$.029 & \textbf{0.848}$\pm$.012 & 0.980$\pm$.003 & 0.933$\pm$.010 \\
        \midrule
        \multicolumn{9}{l}{\textbf{\textit{G-CBM (ours, interpretable concept-graph model)}}} \\
        G-CBM ResNet-50      & \textbf{0.923}$\pm$.004 & \textbf{0.909}$\pm$.015 & \textbf{0.960}$\pm$.006 & \textbf{0.925}$\pm$.004 & 0.868$\pm$.008 & 0.823$\pm$.019 & 0.983$\pm$.002 & 0.935$\pm$.003 \\
        G-CBM DenseNet-201   & 0.907$\pm$.002 & 0.860$\pm$.007 & 0.937$\pm$.015 & 0.908$\pm$.021 & 0.834$\pm$.015 & 0.786$\pm$.009 & \textbf{0.985}$\pm$.001 & \textbf{0.949}$\pm$.006 \\
        G-CBM MobileNet-V2   & 0.892$\pm$.003 & 0.865$\pm$.009 & 0.913$\pm$.006 & 0.889$\pm$.001 & 0.830$\pm$.005 & 0.791$\pm$.004 & 0.956$\pm$.002 & 0.908$\pm$.006 \\
        \midrule
        % \multicolumn{9}{l}{\textit{G-CBM ablations (classifier head variants)}} \\
        MLP-CBM (ours) ResNet-50 &
        0.856$\pm$.002 & 0.799$\pm$.006 &
        0.897$\pm$.006 & 0.854$\pm$.000 &
        0.810$\pm$.008 & 0.762$\pm$.005 &
        0.977$\pm$.000 & 0.925$\pm$.000 \\
        Linear-CBM (ours) ResNet-50 &
        0.850$\pm$.000 & 0.768$\pm$.001 &
        0.887$\pm$.059 & 0.866$\pm$.020 &
        0.810$\pm$.001 & 0.724$\pm$.009 &
        0.971$\pm$.002 & 0.910$\pm$.004 \\
        \bottomrule
    \end{tabular}}
    \caption{Test-set classification performance (AUC and F1, mean\,$\pm$\,std, $n=3$).
    G-CBM models use the per-dataset optimal $\tau^*$ from
    Table~\ref{tab:threshold_graph_summary}. \textbf{Bold} = best value
    per dataset/metric across both baseline and G-CBM groups. MLP-CBM and Linear-CBM replace the GAT with MLP and linear classifiers, respectively.}
    \label{tab:perf}
    \vspace{-3mm}
\end{table*}
\vspace{-18pt}
\subsection{Classification Performance}
\label{sec:perf}
Table~\ref{tab:perf} compares G-CBM with CNN baselines across all four datasets and three backbones, and with CBM (linear) and CBM (MLP) on all four datasets using ResNet-50.
G-CBM (ResNet-50) improves average AUC over the ResNet-50 baseline by
3.7\% relative, and beats the strongest CNN on HAM10000 (0.923 vs.\
0.912) and ImageNet (0.983 vs.\ 0.980); with DenseNet-201, G-CBM
achieves the highest ImageNet AUC overall (0.985). On PH2, G-CBM (ResNet-50,
$\tau^*=0.5$) reaches AUC\,=\,0.960\,$\pm$\,0.006 with
$\bar{K}_{\tau^*}=2.0$ of $r=10$ active concept nodes
(Table~\ref{tab:budget}), showing that selective high-confidence concept
evidence can improve accuracy. Replacing the GAT classifier with linear or MLP classifiers results in suboptimal performance, suggesting that the GAT improves predictive accuracy.

\vspace{-6pt}

\subsection{Comparison with Supervised Methods}
\label{sec:sota}
Table~\ref{tab:sota} compares G-CBM with state-of-the-art supervised concept-based
approaches on Derm7pt and PH2. Without concept annotations, G-CBM
(ResNet-50) achieves 0.868 AUC on Derm7pt, surpassing PCBM, PCBM-h, CBE, and both MICA variants, and trailing only CAW. On PH2, G-CBM reaches 0.960 AUC and 0.925 F1, outperforming~\cite{sarkar2021frameworklearningantehocexplainable}, PCBM, and
PCBM-h, while remaining below CBE and MICA variants.
\vspace{-9pt}
\begin{table}[ht]
    \resizebox{\columnwidth}{!}{%
    \begin{tabular}{lrrrr}
        \toprule
        & \multicolumn{2}{c}{Derm7pt} & \multicolumn{2}{c}{PH2} \\
        \cmidrule(lr){2-3}\cmidrule(lr){4-5}
        Method & AUC & F1 & AUC & F1 \\
        \midrule
        Sarkar~et~al.~\shortcite{sarkar2021frameworklearningantehocexplainable}   & 0.762$\pm$.021 & 0.668$\pm$.012 & 0.793$\pm$.006 & 0.797$\pm$.021 \\
        PCBM~\cite{Yuksekgonul2022}           & 0.730$\pm$.014 & 0.710$\pm$.014 & 0.783$\pm$.012 & 0.815$\pm$.026 \\
        PCBM-h~\cite{Yuksekgonul2022}         & 0.833$\pm$.011 & 0.745$\pm$.014 & 0.923$\pm$.015 & 0.833$\pm$.026 \\
        CBE~\cite{patrício2023coherentconceptbasedexplanationsmedical}                & 0.766$\pm$.004 & 0.781$\pm$.004 & 0.976$\pm$.000 & 0.939$\pm$.000 \\
        MICA~(w/\,bot)~\cite{bie2024micaexplainableskinlesion}         & 0.841$\pm$.011 & 0.781$\pm$.012 & 0.977$\pm$.012 & 0.944$\pm$.015 \\
        MICA~(w/o\,bot)~\cite{bie2024micaexplainableskinlesion}        & 0.856$\pm$.011 & 0.794$\pm$.012 & \textbf{0.982}$\pm$.014 & \textbf{0.953}$\pm$.012 \\
        CAW~\cite{hou2024conceptattentionwhiteninginterpretableskin}   & \textbf{0.886}$\pm$.001 & 0.813$\pm$.009 & ---  & ---  \\
        \midrule
        G-CBM (ours)$^\dagger$            & 0.868$\pm$.008 & \textbf{0.823}$\pm$.019 & 0.960$\pm$.006 & 0.925$\pm$.004 \\
        \bottomrule
    \end{tabular}}
    \caption{Comparison with supervised concept-based methods on
    Derm7pt and PH2 (ResNet-50). Prior-work results
    (mean\,$\pm$\,std) are reproduced from the original papers;
    $\dagger$ denotes our results at the per-dataset optimal $\tau^*$
    (Table~\ref{tab:threshold_graph_summary}), mean\,$\pm$\,std over
    $n{=}3$ seeds. G-CBM is the only method that requires no concept
    annotations. \textbf{Bold} = best per dataset/metric.}
    \label{tab:sota}
    \vspace{-3mm}
\end{table}
\vspace{-10pt}
\subsection{Concept Selectivity}
\label{sec:sparsity-test}
We measure how \emph{selectively} G-CBM uses its concept basis: among the $r$ available nodes, how many actually contribute to each prediction ($\bar{K}_\tau$, Def.~\ref{def:active-count}). Table~\ref{tab:budget} reports
test-set results at the calibrated $\tau^*$ (Sec.~\ref{sec:budget}). On PH2, F1 \textit{improves} from 0.887 at $\tau=0$ to 0.925 at $\tau^*=0.5$ while $\bar{K}_{\tau^*}$ drops from 10 to 2.0; on HAM10000, peak F1 of 0.909 is reached at $\tau^*=0.2$ with only 3.8 of 9 nodes active. F1 alone can be misleading under class imbalance, but AUC at $\tau^*$ (0.923--0.983 across
datasets) confirms the gain is genuine. G-CBM's predictions are thus anchored in a \emph{selective, high-confidence subset of concept evidence}---on PH2 and HAM10000, 20--42\% of available concepts suffice; on Derm7pt and ImageNet, more concepts remain active, reflecting greater visual heterogeneity rather than diffuse averaging.
\begin{table}[t]
    \resizebox{\columnwidth}{!}{%
    \begin{tabular}{lrrrrrr}
        \toprule
        Dataset & $r$ & F1 ($\tau\!=\!0$) & $\tau^*$ & F1 ($\tau^*$) & AUC ($\tau^*$) & $\bar{K}_{\tau^*}$ \\
        \midrule
        HAM10000 & 9  & 0.885$\pm$.005 & 0.2 & \textbf{0.909}$\pm$.015 & 0.923$\pm$.004 & 3.8 \\
        PH2      & 10 & 0.887$\pm$.002 & 0.5 & \textbf{0.925}$\pm$.004 & \textbf{0.960}$\pm$.006 & 2.0 \\
        Derm7pt  & 12 & 0.806$\pm$.019 & 0.1 & \textbf{0.823}$\pm$.019 & \textbf{0.868}$\pm$.008 & 9.0 \\
        ImageNet & 8  & 0.944$\pm$.006 & 0.1 & 0.935$\pm$.003 & 0.983$\pm$.002 & 6.4 \\
        \bottomrule
    \end{tabular}}
    \caption{Test-set concept selectivity (ResNet-50). $r$: total concept
    nodes; $\bar{K}_{\tau^*}$: mean active nodes at $\tau^*$.
    F1 at $\tau=0$ is the unfiltered baseline; models are retrained with
    $\tau^*$ applied throughout. AUC confirms the gain is not an
    F1-only artefact on imbalanced datasets.}
    \label{tab:budget}
    \vspace{-3mm}
\end{table}

\vspace{-1mm}
\subsection{Faithfulness Analysis}
\label{sec:fidelity}
We assess whether the GAT's gradient-ranked concept importance forms a
\emph{faithful} explanatory signal via deletion/insertion analysis, a
standard perturbation-based evaluation in
XAI~\cite{samek2016evaluating,Petsiuk2018,fel2023holistic,achtibat2023attribution,kadir2023evaluation}.
For each test image we compare two rankings of the $r$ concept nodes:
\textit{Most-Relevant-First (MRF)}, nodes ranked by $S_C(v,c^*)$
(Eq.~\ref{eq:concept-score}), and \textit{Random}, a uniformly sampled
permutation serving as a no-information control. Deletion progressively
zeroes the top-$\lfloor fr \rfloor$ node features for
$f \in \{0,0.1,\ldots,1\}$; insertion restores them from an all-zero
state. We summarise each 11-point curve by trapezoidal AUC:
AUC$_\text{del}$ ($\downarrow$) and AUC$_\text{ins}$ ($\uparrow$). Across all four datasets (Table~\ref{tab:fidelity}), MRF outperforms
the random baseline with AUC$_\text{del}$ lower by 10--16 points and
AUC$_\text{ins}$ higher by 5--10 points, confirming that gradient-ranked
concept importance carries genuine explanatory signal.
\vspace{-6pt}
\begin{table}[ht]
    \resizebox{\columnwidth}{!}{%
    \begin{tabular}{lrrrr}
        \toprule
        & \multicolumn{2}{c}{AUC$_\text{del}$ $\downarrow$}
        & \multicolumn{2}{c}{AUC$_\text{ins}$ $\uparrow$} \\
        \cmidrule(lr){2-3}\cmidrule(lr){4-5}
        Dataset & MRF & Random & MRF & Random \\
        \midrule
        HAM10000 & \textbf{0.767}$\pm$.171 & 0.872$\pm$.154 & \textbf{0.922}$\pm$.126 & 0.871$\pm$.158 \\
        PH2        & \textbf{0.742}$\pm$.260 & 0.899$\pm$.147 & \textbf{0.967}$\pm$.037 & 0.868$\pm$.175 \\
        Derm7pt   & \textbf{0.611}$\pm$.064 & 0.770$\pm$.124 & \textbf{0.861}$\pm$.132 & 0.766$\pm$.128 \\
        ImageNet  & \textbf{0.706}$\pm$.238 & 0.855$\pm$.154 & \textbf{0.941}$\pm$.085 & 0.863$\pm$.151 \\
        \bottomrule
    \end{tabular}}
    \caption{Faithfulness of concept-node importance rankings via
    deletion/insertion analysis. AUC computed by trapezoidal integration
    over $f \in [0,1]$ (11 points); $\pm$ = per-image std. MRF = Most-Relevant-First (gradient-ranked). Lower AUC$_\text{del}$
    and higher AUC$_\text{ins}$ indicate more faithful explanations.}
    \label{tab:fidelity}
       \vspace{-4mm}
\end{table}

\vspace{-5pt}

\subsection{Spatial Interpretability}
\label{sec:qual}

\begin{figure*}[t]
    \centering
    \includegraphics[width=\linewidth]{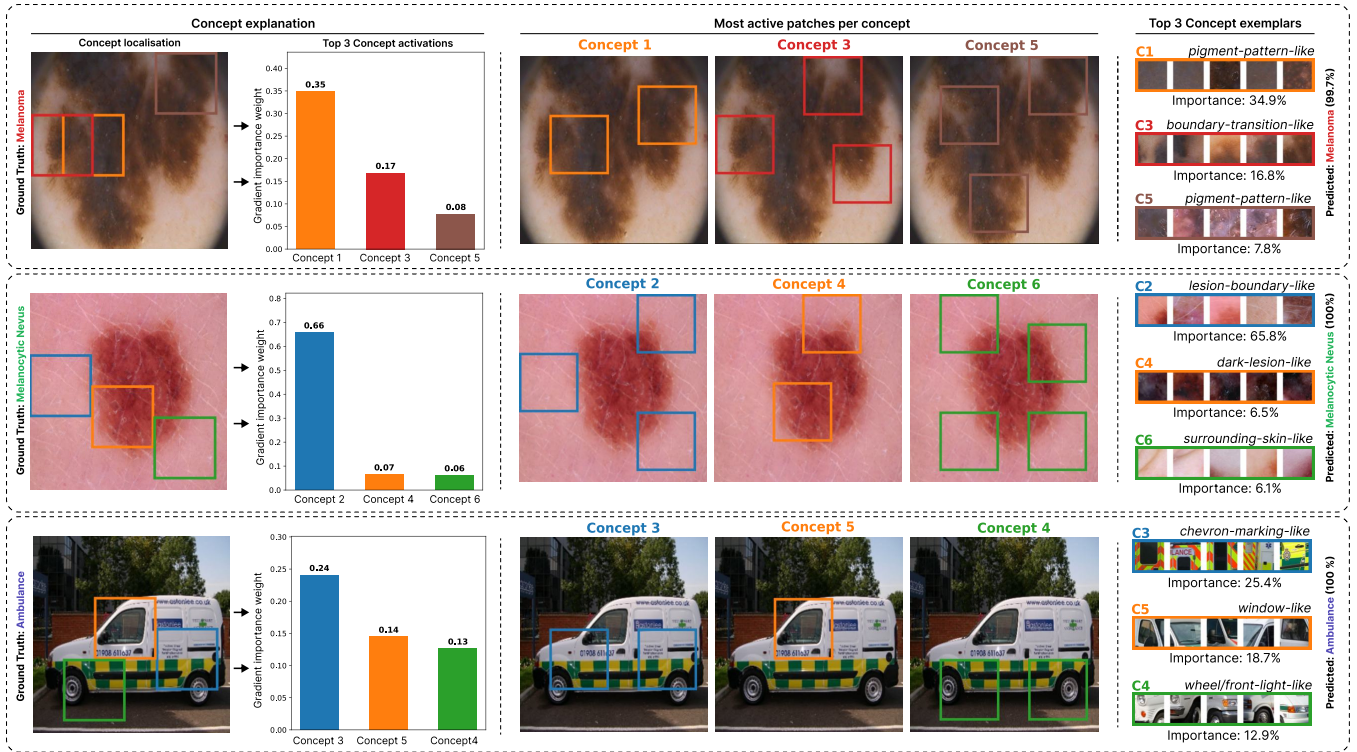}
    \caption{Concept-level spatial explanations on PH2, HAM10000, and
ImageNet, illustrating G-CBM's three interpretability outputs:
\emph{concept selection} (top-3 concept nodes), \emph{concept
grounding} (coloured bounding boxes), and \emph{importance scores}
($S_C(v,c^*)$, Eq.~\ref{eq:concept-score}, normalised over the top-3).
Each row also shows per-concept activation overlays and exemplar
patches retrieved by highest NMF score $s_{jv}$.
\textit{Row 1 (PH2 -- Melanoma, 99.7\%)}: top concept node (orange,
$S_C{=}0.35$) on the atypical pigment network at the lesion centre.
\textit{Row 2 (HAM10000 -- Nevus, 100\%)}: dominant concept node
(blue, $S_C{=}0.66$) along the lesion boundary and adjacent skin.
\textit{Row 3 (ImageNet -- Ambulance, 100\%)}: top concept node on
chevron-style markings; additional nodes cover side window/signage and
lower vehicle structure. ``-like'' descriptors are post-hoc summaries
of retrieved exemplars, not training supervision.}
    \vspace{-4mm}
    \label{fig:spatial}
\end{figure*}

Figure~\ref{fig:spatial} shows concept-level explanations for three
correctly classified images spanning all three data domains: a PH2
melanoma, a HAM10000 melanocytic nevus, and an ImageNet ambulance. In each case, 
the model identifies a small set of localized concepts, assigns them class-relevant 
importance scores, and retrieves exemplar patches that support their interpretation. 
The examples show that concept evidence remains spatially concentrated and semantically 
coherent across domains: the melanoma case highlights the lesion core, the nevus case emphasizes 
the lesion boundary, and the ambulance case separates vehicle markings from surrounding
structural regions. This consistency indicates that G-CBM learns concept-level evidence 
that is both localizable and transferable beyond dermoscopy.
\vspace{-5pt}
\subsection{Patch-Stride Ablation}
\label{sec:patch}
We systematically investigate the effect of the patch extraction hyperparameters
via a grid search over patch sizes $p \in \{48, 56, 64, 70, 96, 112\}$ and
stride ratios $\rho \in \{0.25, 0.35, 0.50, 0.65, 0.80\}$ for G-CBM
(ResNet-50) on all four datasets. Table~\ref{tab:patch} reports selected
results with a standard ResNet-50 CNN baseline as reference. The setting
$p\!=\!70$, $\rho\!=\!0.5$ outperforms the CNN baseline on all three metrics
(AUC, F1, Accuracy) in three of four datasets (HAM10000, Derm7pt, ImageNet;
PH2 is the exception at this stage) and was selected as the universal
default.

\begin{table}[t]
    \small
    \begin{tabular}{llrrrrl}
        \toprule
        Model & $p$/$\rho$ & HAM & PH2 & D7 & IN & $\checkmark$ \\
        \midrule
        Baseline & --- & 0.891 & 0.903 & 0.828 & 0.980 & \\
        \midrule
        G-CBM & 48 / 0.5           & \textbf{0.937} & \textbf{0.930} & 0.835 & 0.972 & \\
        G-CBM & 70 / 0.5           & 0.928 & 0.910 & \textbf{0.863} & 0.984 & $\checkmark$ \\
        G-CBM & 96 / 0.65          & 0.920 & 0.920 & 0.848 & \textbf{0.991} & $\checkmark$ \\
        G-CBM & 112 / 0.5          & 0.924 & 0.910 & 0.845 & 0.989 & \\
        \bottomrule
    \end{tabular}
    \caption{Patch-stride ablation (AUC on test split). ResNet-50 CNN is
    the baseline. HAM\,=\,HAM10000; D7\,=\,Derm7pt; IN\,=\,ImageNet. \textbf{Bold} marks the best G-CBM value among the selected patch-stride settings for each dataset. $\checkmark$ indicates all three
    metrics (AUC, F1, Acc) exceed the CNN baseline on at least 3 of 4 datasets.} 
    \label{tab:patch}
    \vspace{-3mm}
\end{table}
\vspace{-10pt}
\section{Conclusion}
We introduced G-CBM, an intrinsically interpretable framework for
visual classification that combines unsupervised concept discovery
(via NMF) with graph attention over a per-image concept-graph. G-CBM jointly
produces \emph{concept selection} (\emph{what}), \emph{concept grounding}
(\emph{where}), and \emph{importance scores} (\emph{how much}) 
within the prediction pipeline, without requiring supervised concept annotations, LLM/CLIP-assisted discovery, or post-hoc analysis on a pretrained model.

G-CBM improves average AUC by 3.7\% relative over ResNet-50 across four datasets. The concept filtering threshold $\tau$
yields selective, high-confidence explanations: on PH2, only two of ten
concept nodes are active on average. Deletion/insertion analysis with random-ordering controls confirms that concept
importance carries a genuine explanatory signal.
G-CBM's patch representation depends on sufficient resolution and can fragment structures at patch boundaries; overlapping patches partly mitigate this, and segmentation-guided extraction is a natural next step. Overall, G-CBM shows that visual classifiers can reason through localised, reusable
concepts without sacrificing predictive performance.

\section*{Acknowledgments}
This work was partially funded by the Federal Ministry of Research, Technology and Space (BMFTR) under grant number 16IW23002 (No-IDLE), grant number 16IW24006 (NoIDLEChatGPT), and grant number 16IW26002 (Amenable); the Lower Saxony Ministry of Science and Culture (MWK) in the zukunft.niedersachsen program; and the Endowed Chair of AAI at the University of Oldenburg.

\bibliographystyle{named}
\bibliography{ijcai26}

\end{document}